%% file: CMR.tex
\documentclass{article} % For LaTeX2e
\usepackage{iclr2023_conference,times,url,booktabs,microtype,graphicx,subfigure,bbm,verbatim,amsthm,amssymb,amsmath,enumitem,pifont,makecell,multirow,mdframed,xcolor,colortbl,rotating,wrapfig,resizegather,cancel}
\usepackage[export]{adjustbox}

\usepackage{pifont}
\usepackage{tikz}
\newcommand*\circled[1]{\tikz[baseline=(char.base)]{
            \node[shape=circle,draw,inner sep=0.4pt] (char) {#1};}}

\definecolor{CColor}{rgb}{0.01,0.31,0.59}
\definecolor{GGray}{rgb}{0.80,0.90,1}
\definecolor{Shady}{rgb}{0.9,0.9,0.9}
\definecolor{kaistblue}{RGB}{20,135,200}
\definecolor{kaistdarkblue}{RGB}{0,65,145}
\definecolor{urbanablue}{RGB}{19,41,75}
\definecolor{urbanaorange}{RGB}{232,74,39}
\definecolor{drp}{rgb}{0.53,0.15,0.34}
\usepackage[plainpages=false,pdfpagelabels,colorlinks=true,linkcolor=CColor,citecolor=CColor,urlcolor=CColor]{hyperref}
\input{math_commands.tex}

\newcolumntype{P}[1]{>{\centering\arraybackslash}p{#1}}

\title{Sparsity May Cry: Let Us Fail (Current) Sparse Neural Networks Together!}  

\author{Shiwei Liu\textsuperscript{1}$^*$, 
  Tianlong Chen\textsuperscript{1}$^*$, 
  Zhenyu Zhang\textsuperscript{1},
  Xuxi Chen\textsuperscript{1},
  Tianjin Huang\textsuperscript{2},\\
  \textbf{Ajay Jaiswal\textsuperscript{1}},
  \textbf{Zhangyang Wang\textsuperscript{1}}\\
  {\textsuperscript{1}University of Texas at Austin \,\,} 
    {\textsuperscript{2}Eindhoven University of Technology}
  \\
  \small{\texttt{shiwei.liu@austin.utexas.edu}}; 
  \small{\,\,\,\texttt{t.huang@tue.nl}}\\
  \small{\texttt{\{tianlong.chen,zhenyu.zhang,xxchen,ajayjaiswal,atlaswang\}@utexas.edu}}\\
}

\iclrfinalcopy % Uncomment for camera-ready version, but NOT for submission.
\begin{document}

\maketitle

\def\thefootnote{*}\footnotetext{These authors contributed equally to this work.}\def\thefootnote{\arabic{footnote}}

\begin{abstract}

Sparse Neural Networks (SNNs) have received voluminous attention predominantly due to growing computational and memory footprints of consistently exploding parameter count in large-scale models. Similar to their dense counterparts, recent SNNs generalize just as well and are equipped with numerous favorable benefits (e.g., low complexity, high scalability, and robustness), sometimes even better than the original dense networks. As research
effort is focused on developing increasingly sophisticated sparse algorithms, it is startling that a \textit{comprehensive benchmark to evaluate the effectiveness} of these algorithms has been highly overlooked. In absence of a carefully crafted evaluation benchmark, most if not all, sparse algorithms are evaluated against fairly simple and naive tasks (eg. CIFAR-10/100, ImageNet, GLUE, etc.), which can potentially \textit{camouflage} many advantages as well unexpected predicaments of SNNs. In pursuit of a more general evaluation and unveiling the true potential of sparse algorithms, we introduce ``\textbf{Sparsity May Cry}'' \textbf{Benchmark (SMC-Bench)}, a collection of carefully-curated 4 diverse tasks with 10 datasets, that accounts for capturing a wide range of domain-specific and sophisticated knowledge. Our systemic evaluation of the most representative sparse algorithms reveals an important obscured observation: \textit{the state-of-the-art magnitude- and/or gradient-based sparse algorithms seemingly fail to perform on SMC-Bench} when applied out-of-the-box, sometimes at significantly trivial sparsity as low as $5\%$. The observations seek the immediate attention of the sparsity research community to reconsider the highly proclaimed benefits of SNNs. We further conduct a thorough investigation into the reasons for the failure of common SNNs. Our analysis points out that such failure is intimately related to the ``lazy regime” of large model training,  which hints us with stronger pruning recipes that alleviate the failure on SMC-Bench (though still more or less suffering). By incorporating these well-thought and diverse tasks, SMC-Bench is designed to favor and encourage the development of more scalable and generalizable sparse algorithms. We open-source SMC-Bench to assist researchers in building next-generation sparse algorithms that scale and generalize: \url{https://github.com/VITA-Group/SMC-Bench}.

% -------------------------------

\end{abstract}
\section{Introduction}
\vspace{-0.5em}
Sparse Neural Networks (SNNs) are no stranger to the deep learning community \citep{liu2023ten}, but recently they have received stupendous attention in the era of transformers (eg. BERT, GPT, ViT, CLIP, etc.), when the parameter count is frequently measured in billions rather than millions. Due to the consistent efforts of sparsity researchers, SNNs have ushered enormous breakthroughs and can generalize just as well as original dense networks, and it is feasible to procure them after training~\citep{frankle2018lottery,sanh2020movement,chen2020lottery,frankle2020early}, during training~\citep{zhu2017prune,gale2019state,liu2021sparse}, and even before training~\citep{mocanu2018scalable,lee2018snip,liu2022the} their dense counterparts using pruning. Apart from well-known efficiency benefits, surprisingly, SNNs also enjoy auxiliary benefits such as adversarial robustness~\citep{guo2018sparse,ozdenizci2021training,chensparsity}, out-of-distribution generalization~\citep{zhang2021can,diffenderfer2021winning}, and uncertainty estimation~\citep{liu2021deep}, etc. Despite the multi-dimensional success of numerous sparse algorithms, startlingly, our extensive survey across over 100 recent SNN papers within \texttt{2015-2022} unveils multiple daunting issues regarding evaluation datasets and protocols blindly followed within the sparsity community, that may significantly impede future progress if left unacknowledged.

\textbf{Issues with current evaluation paradigm:}  \textit{Firstly}, the vast majority of current work on SNNs is \textit{narrowly evaluated}, i.e., only targeting a single or a few tasks (usually on image classification and sometimes on language understanding) on which SNNs have already proven their proficiency~\citep{gale2019state,frankle2018lottery}. Surprisingly, 79 papers out of our carefully selected 100 papers on SNNs, evaluate sparse models \textit{merely} on a single task, where 72 out of them evaluate image classification. \textit{Secondly}, people are obsessed with evaluating SNNs on well-understood datasets, including but not limited to MNIST~\citep{lecun1998mnist} (26 papers), CIFAR-10/100~\citep{krizhevsky2009learning} (59 and 37 papers,  respectively), ImageNet~\citep{deng2009imagenet} (62 papers), and GLUE~\citep{wang2018glue} (9 papers), where deep neural networks have already exceeded the human-equivalent performance (refer to Appendix~\ref{app:summary_100papers} for more details). For instance, even though ImageNet has been considered a rather challenging task over years, very high accuracy ($>$90\%) has been reported many times~\citep{yu2022coca,wortsman2022model,zhai2022scaling}. Such relatively restricted evaluations with ``nearly saturated performance" limit the scope of sparse neural networks and are potentially ill-suited to identify new and unexpected capabilities of SNNs. 

% Obsessing evaluating sparse neural networks on fairly simple dataset will dilute the performance difference of newly proposed sparsification methods, preventing the quantitative and further the qualitative improvements of sparse neural networks.

% Therefore, addressing the above-mentioned limitations of current benchmarks is urgently required and is beneficial for the machine learning community, due to the critical role of SNNs in industry and academia. In this work, we assemble a large-scale, difficult and diverse benchmark for sparse neural networks - ``\textbf{Sparsity May Cry Benchmark''} (or briefly \textbf{SMC-Bench}). Specifically, we consider a broad set of tasks including complex reasoning, multilingual translation, and protein prediction, whose content spans multiple domains, requiring a vast amount of commonsense knowledge, solid mathematical and scientific background to solve. We broadly measure the performance of the most  state-of-the-art sparse subnetworks produced by the off-the-shelf SOTA pruning and sparse training approaches on SMC-Bench across sparsities, to analyze whether the behavior of SNNs is significantly distinguishable from the dense ones. In our analysis, we do not focus on the performance of SNNs on any single task, but rather on how the performance changes across a wide range of tasks. Our contributions are summarized below:

Addressing the aforementioned limitations of current SNN evaluation protocols is a pressing need for the community. To this end, we assemble a large-scale, fairly arduous, and diverse benchmark for sparse neural networks - ``\textbf{Sparsity May Cry}'' \textbf{Benchmark} (or briefly \textbf{SMC-Bench}). Specifically, we consider a broad set of tasks including \textit{complex reasoning, multilingual translation, and protein prediction}, whose content spans multiple domains. Those tasks require a vast amount of commonsense knowledge, solid mathematical and scientific backgrounds to solve even for humans. Note that none of the datasets
in SMC-Bench was created from scratch for the benchmark, we rely on pre-existing datasets as they have been agreed by researchers as challenging, interesting, and of high practical value. We rigorously measure the performance of state-of-the-art (SOTA) pruning and sparse training approaches (in their most common, basic settings) on SMC-Bench, to understand the potential of SNNs to scale and generalize. Our key observations and contributions can be unfolded as:

\begin{itemize}
    \item We present ``\textbf{Sparsity May Cry}'' Benchmark, to \underline{\textbf{re-define}} the evaluation protocols for sparse neural networks and facilitate a comprehensive assessment of SOTA sparse algorithms. The premise of SMC-bench is to develop a suite of large-scale, challenging, realistic, and diverse tasks and datasets that can empower the rigorous advancements in the community. 
    
    \item SMC-Bench unveils a critical and startling observation - all of the SOTA sparse algorithms seem to fail on SMC-Bench ``out-of-the-box", sometimes at significantly trivial sparsity \textit{e.g., $5\%$}. Note that the failure does not appear specific to one sparsification approach but unanimously across all approaches we evaluated. This observation \textit{alarmingly} demands the attention of the sparsity community to reconsider the highly proclaimed benefits of SNNs.
    
    \item We conduct extensive experiments across representative SNNs produced by various SOTA pruning and sparse training approaches on SMC-Bench, and we summarize our findings: \ding{182} Model prunability is intimately related to task difficulty: models trained on difficult tasks suffer more from pruning compared to easier tasks; \ding{183} The success of before-training sparsification (sparse training or pruning at initialization) is hard to generalize in more complex scenarios; \ding{184} Iterative magnitude pruning (IMP) does not necessarily generalize better than one-shot pruning (OMP) or during-training pruning; \ding{185} Despite performance difference, different magnitude-based pruning approaches lead to extremely similar layerwise sprasities. 

\item We further carry out a comprehensive investigation into the potential causes of SNN failures on SMC-Bench. Our analysis suggests that the failure of the existing sparse algorithms might be due to the ``lazy regime'' dynamics emerging in sufficiently overparameterized models~\citep{chizat2019lazy,malladi2022kernel}. Inspired by this finding,  we hypothesize and confirm that the second-order pruning approaches, i.e., oBERT~\citep{kurtic2022optimal}, are more reliable pruning approaches for SMC-Bench, which yield relatively more promising performance on SMC-Bench in Appendix~\ref{app:root_cause}.

\end{itemize}

\vspace{-0.5em}
\section{Related Work}
\vspace{-0.5em}

\subsection{Advances in Sparse Neural Networks}
\vspace{-2mm}
\textbf{Post-Training.} SNNs refer to a neural network where a certain portion of its components (e.g., weights, neurons, filters, and attention heads) have exactly zero values. The initial purpose of SNNs is retrospectively to accelerate model at inference time (a.k.a., post-training sparsification;~\citet{mozer1989using,lecun1990optimal}). Thanks to the over-parameterization property of deep neural networks, we can dramatically prune deep neural networks to smaller sizes with marginal loss of performance. Post-training sparsification has been well studied and results in various mature criteria that can be generally categorized into zero-order methods (magnitude-based;~\citet{han2015deep}), first-order methods (gradient-based;~\citet{molchanov2016pruning,sanh2020movement,jiang2021towards}), and second-order methods (Hessian-based;~\citet{lecun1990optimal,hassibi1992second,dong2017learning}). Second-order sparsification usually achieves higher performance than the other two but is also more expensive due to the full Hessian calculation. Fortunately, many approaches have been proposed to efficiently approximate Hessian~\citep{zeng2018mlprune,wang2019eigendamage,singh2020woodfisher}. The Lottery Ticket Hypothesis (LTH) adopts iterative magnitude pruning (IMP) on fully trained networks and finds a subnetwork at initialization that can be re-trained in isolation to match the original dense networks.~\citet{Renda2020Comparing} further found that instead of re-training with the initial weights, re-training with the final weights achieves better performance. With the rise of large language models (LLMs), newer post-training pruning methods have emerged which aim to improve the affordability of these models~\citep{sanh2020movement,chen2020lottery,zafrir2021prune,kurtic2022optimal,xu2021rethinking,lagunas2021block,zhang2022platon,frantar2021m}.

\textbf{During-Training.} During-training sparsification~\citep{finnoff1993improving} is a cheaper option, compared to sparsify a fully converged model. Approaches of during-training sparsification usually train a dense network for some time and then gradually sparsify the network with some schedules and end up with a sparse model with target sparsities.~\citet{zhu2017prune,gale2019state,Lin2020Dynamic,liu2021sparse} are highlight approaches that gradually prune networks during training and meanwhile allow the pruned weights to be reactivated in case of inaccurate pruning. Another direction of during-training sparsification is adding sparsifying penalties such as (grouped) $L_0$ and $L_1$ norm to the loss function, which will punish the unimportant weights to zero, leading to sparse weights~\citep{louizos2017learning,luo2020autopruner,savarese2020winning}.

\textbf{Prior-Training.} Recently, foundation models~\citep{brown2020language,chowdhery2022palm,ramesh2022hierarchical} have demonstrated promising quantitative improvement and new qualitative capabilities with increasing scale~\citep{zhang2020you}. Along with the scaling of model size and data size, the training resources of these foundation models also get outrageous. To accelerate training, we need to sparsify models before training. LTH unveils the possibility to find SNNs at initialization that can match their dense counterparts, even though it uses post-training pruning to find them. At the same time, sparse training~\citep{mocanu2018scalable,mostafa2019parameter,dettmers2019sparse,evci2020rigging,liu2021we,schwarz2021powerpropagation} was proposed that can train a randomly-initialized sparse neural network from scratch while dynamically optimizing the sparse connectivity with promising performance. Instead of randomly initializing sparse networks, one iteration~\citep{lee2018snip,Wang2020Picking} or a few iterations~\citep{tanaka2020pruning,de2020progressive} of training can be utilized to guide the search for sparse networks before training. 

\vspace{-2mm}
\subsection{Benchmarking in Sparse Neural Networks}
\vspace{-2mm}

\citet{gale2019state} rigorously evaluated variational dropout~\citep{molchanov2017variational}, $l_0$ regularizaion~\citep{louizos2017learning}, and GMP~\citep{zhu2017prune} on two large-scale tasks. They demonstrated that the appealing performance advantages of variational dropout and $l_0$ regularization cannot generalize to large-scale tasks whereas simple magnitude pruning performs surprisingly well.~\citet{liu2018rethinking} examined two pipelines: training from scratch and fine-tuning, concluding that fine-tuning a pruned model only gives comparable or worse performance than training from scratch.~\citet{blalock2020state} provided a comprehensive literature review on SNNs and found that pruning papers rarely make direct and controlled comparisons.~\citet{frankle2021pruning} assessed the efficacy of various pruning at initialization approaches and attribute their inferior performance
to their insensitivity to weight shuffling and re-initialization.~\citet{liu2022the} re-evaluated the performance of various random pruning before training and found that sparsely training a randomly pruned network from scratch can surprisingly match the performance of its dense equivalent. These papers shed light on the behavior of SNNs and discover important research problems for future work.

\vspace{-0.5em}
\section{SMC-Bench}
\vspace{-2mm}
SMC-Bench is crafted for evaluating if all proclaimed benefits of SNNs can ``scale and generalize”. It consists of 4 diverse and difficult tasks, including commonsense reasoning, arithmetic reasoning, multilingual translation, and protein prediction, with 10 datasets collected from prior work and open-source GitHub repositories. To investigate if there is a strong correlation between model prunability and task difficulty, we choose multiple datasets with different degrees of difficulty.

\vspace{-1mm}
\subsection{Commonsense Reasoning}
\vspace{-2mm}
Commonsense reasoning task asks commonsense questions about the world involving
complex semantics that often require rich common sense and background knowledge. We consider three commonly used datasets for commonsense reasoning. (1)~\textbf{RACE}~\citep{lai2017large} contains near 28,000 passages and 100,000 questions collected from the English exams for Chinese students in middle (RACE-M) and high school (RACE-H). (2)~\textbf{WinoGrande}~\citep{sakaguchi2021winogrande} is a modified version of the Winograd Schema Challenge (WSC) benchmark~\citep{levesque2012winograd} with enhanced scale and hardness, containing 44k problems. (3)~\textbf{Commonsense QA (CSQA)}~\citep{talmor2018commonsenseqa} is a challenging dataset containing 12,247 multiple-choice questions where one source concept and three target concepts are first extracted from ConceptNet~\citep{speer2017conceptnet} based on which the Crowd-works are asked to author multiple-choice questions with two additional distractors. In general, CSQA is harder than WinoGrande and RACE, with ceiling human performance of 89\%, 94\%, and 95\%, respectively.

\vspace{-2mm}
\subsection{Arithmetic Reasoning}
\vspace{-0.5em}
Arithmetic reasoning poses a question of a math problem and the model is asked to generate a mathematical equation that can be evaluated to get the answer. We consider the following three math word problem datasets: (1) the widely used \textbf{MAWPS}
benchmark~\citep{koncel2016mawps} composed of 2,373 problems; (2) the arithmetic subset of ASDiv~\citep{miao2021diverse} - \textbf{ASDiv-A} with 1,218 math problems; (3) the more challenging \textbf{SVAMP}~\citep{patel2021nlp} dataset which is created by applying complex types of variations to the samples from ASDiv-A. The task difficulty monotonically increases from MAWPS to ASDiv-A, and to SVAMP.

\vspace{-2mm}
\subsection{Protein Thermostability Prediction}
\vspace{-0.5em}
Maintaining a stable 3D structure is an essential pre-condition for protein to function correctly in biological phenomena. Numerous efforts are devoted to modeling and predicting protein's stability against pH, salinity, and temperature. We consider the tasks of protein thermostability prediction on two representative datasets: (1) \textbf{HotProtein}~\citep{hp} is recently proposed as a large-scale, standardized protein benchmark with organism-level temperature annotations, which contains $182$K protein sequences and $3$K folded structure from $230$ different species. Three dataset variants, \textbf{HP-S}, \textbf{HP-S$^2$C$5$}, and \textbf{HP-S$^2$C$2$}, are adopted to examine sequence- and structure-based methods, respectively. \textbf{HP-S} has \{$6,390$, $3,4946$, $30,333$, $79,087$, $31,549$\} protein sequences from five categories of \{\textit{Cryophilic}, \textit{Psychrophilic}, \textit{Mesophilic}, \textit{Thermophilic}, \textit{Hyperthermophilic}\}; \textbf{HP-S$^2$C$5$} has both sequences and structures for \{$73$, $387$, $195$, $196$, $189$\}
proteins from the same five classes ordered from \textit{Cryophilic} to \textit{Hyperthermophilic}; \textbf{HP-S$^2$C$2$} has both sequences and structures for \{$1,026$, $939$\} proteins from \{``hot'' ($\ge 45^{\circ}\texttt{C}$), ``cold'' ($< 45^{\circ}\texttt{C}$)\} two classes. (2) \textbf{Meltome Atlas}~\citep{jarzab2020meltome} is another challenging test bed for protein's thermostability. It has \{$7,902$, $15,833$, $10,518$\} protein sequences from three of the five aforementioned classes, from \textit{Mesophilic} to \textit{Hyperthermophilic}. All samples are annotated with their melting temperature. 

\vspace{-2mm}
\subsection{Multilingual Translation}
\vspace{-0.5em}
Multilingual translation processes multiple languages using a single language model and requires the model to have the ability to perform translation across languages. We follow~\citet{liu2020multilingual,tang2020multilingual} and choose 10 English-centric language pairs (Fr, Cs, De, Gu, Ja, My, Ro, Ru, Vi, Zh $\leftrightarrow$ En) from an open source parallel corpus - OPUS~\citep{OPUS}.  We follow~\citet{arivazhagan2019massively} and use pivot data through English to create 3
Many-to-Many multilingual translation fine-tuning settings including $2$-to-$2$ (Fr, Cs), $5$-to-$5$ (Fr, Cs, De, Gi, Ja), and $10$-to-$10$.

\vspace{-0.5em}
\section{Evaluation on SMC-Bench}
\vspace{-0.6em}
\paragraph{Models.} Despite we are aware that performing few-shot prompting on large-scale pre-training language models with billions of parameters is able to solve these tasks~\citep{wei2022chain,srivastava2022beyond}, we choose to focus on  fine-tuning or training with pre-trained mid-scale models with millions of parameters, to improve the accessibility of our Benchmark. Specifically, we choose to fine-tune the popular RoBERTa~\citep{liu2019roberta} for commonsense reasoning; to fine-tune mBART~\citep{liu2020multilingual} for multilingual translation; to train GTS~\citep{xie2019goal} and Graph2Tree~\citep{zhang2020graph} with RoBERTa's pre-trained embedding for arithmetic reasoning; to fine-tune Transformer-based~\citep{vaswani2017attention} for protein property prediction.  See Appendix~\ref{app:summary} for full details.

\textbf{Sparse Neural Networks.} We select the most representative magnitude-
and/or gradient-based approaches where the prune operation is performed before, during, or after training. Formally, given a dense network $\theta_l \in \mathbb{R}^{d_l}$ with a dimension of $d_l$ in each layer $l \in \left\{1,\dots,L\right\}$, pruning generates binary masks $m_l \in \left\{0,1\right\}^{d_l}$ yielding sparse neural networks with sparse weights $\theta_l \odot m_l$. The sparsity level is the fraction of the weights that are zero-valued, calculated as $s = 1- \frac{\sum_{l}{m_l}}{\sum_{l}{d_l}}$. Following a mainstream convention in many sparse training  papers~\citep{frankle2018lottery,gale2019state,evci2020rigging,lee2018snip,liu2021we}, we sparsify most layers in the model including embedding layers and classifier heads, and  we do not apply advanced techniques such as Iterative Learning Rate Rewinding~\citep{Renda2020Comparing} and Knowledge Distillation~\citep{hinton2015distilling} in our main evaluations, even if we observe that they help to alleviate accuracy drops as in Appendix~\ref{app:root_cause}.

$\bullet$ \textit{Lottery Ticket Hypothesis (LTH)}~\citep{frankle2018lottery} is a strong post-training pruning baseline that iteratively adopts magnitude pruning after training to produce binary masks and re-train together with weights from step $t$. We set $t=0$ in this paper, since rewinding to early training does not notably improve the performance of Transformer models (e.g., BERT) for downstream tasks~\citep{chen2020lottery}. The pruning rate of each IMP is set as 20\%.

$\bullet$ \textit{Magnitude After Training} is a strong baseline for prune after training, which has demonstrated strong results in various regimes. After training or fine-tuning models on the specific task, we prune the model with one-shot magnitude pruning and re-train it with the full learning rate schedule from the beginning, dubbed ``OMP (After)'' in our experiments.

$\bullet$ \textit{Random After Training}~\citep{mittal2019studying} is the most naive baseline for post-training pruning. It uniformly samples a random score $s_l \in \text{Uniform}(0, 1)$ for each weight and prunes the weights with the lowest scores. After pruning, we also re-train with the entire training schedule.

$\bullet$ \textit{Gradual Magnitude Pruning (GMP)}~\citep{zhu2017prune,gale2019state} gradually sparsifies networks during training according to a pre-defined sparsification schedule with sorting-based weight thresholding. The starting and the ending iteration of the gradual sparsification process are set as 10\% and 80\% of the entire training iterations. The frequency of sparsification steps is tuned among 500, 1000, and 4000, depending on the specific tasks. While we are aware of the advanced gradual pruning methods - movement pruning~\citep{sanh2020movement}, it usually exceeds GMP only at high sparsities (e.g., $>$90\%), which is interesting but not within the scope of this paper.

$\bullet$ \textit{Magnitude Before Training}~\citep{frankle2021pruning} simply removes weights with the lowest magnitude at initialization. Since we inherit weights from pre-trained models, the initial weights actually refer to the weights that are learned on the pre-trained tasks. We abbreviate this approach to ``OMP (Before)'' as we use one-shot magnitude pruning. 

$\bullet$ \textit{Random Before Training}~\citep{liu2022the} is the most naive baseline for prior-training pruning. We randomly sample scores for each weight and removes the weights with the lowest scores. Different from Random After Training, the pruning operation is performed before fine-tuning.

$\bullet$ \textit{SNIP}~\citep{lee2018snip} is a prior-training pruning technique that globally removes weights with the lowest connection sensitivity score $|g \odot w|$. SNIP is a strong baseline that consistently  performs well among various prior-training approaches~\citep{frankle2021pruning}.
 
$\bullet$ \textit{Rigging the Lottery (RigL)}~\citep{evci2020rigging} is a leading sparse training method that updates the topology of sparse neural networks during training via a prune-and-grow scheme. To evaluate its effectiveness on downstream fine-tuning, we combine RigL with the other three prior-training methods. The update interval of RigL is set the same as the ones used for updating sparsity in GMP, following~\citet{liu2021sparse}.

\vspace{-0.8em}
\subsection{Commonsense Reasoning} 
\vspace{-2mm}
\paragraph{Implementation Details.} We follow the training settings of sequence modeling toolkit Fairseq~\citep{ott2019fairseq} and fine-tune the pre-trained RoBERTa on our datasets with a standard cross-entropy loss. Specifically for each question, we also construct five inputs, one for each of the five candidate answer choices. Each input is constructed by concatenating the question and candidate answer together. We then encode each input and pass the resulting ``[CLS]" representations through a classifier to predict the correct answer. All models are trained with the Adam~\citep{kingma2014adam} optimizer with a learning rate of $1\times 10^{-5}$ using an A100 GPU. For CSQA, we train the model for 3000 steps with a linear warmup of 150 steps and a batch size of $16$. The dropout rate is set as 0.1. This gives us a test accuracy of $77.3\%$ with dense RoBERTa. For RACE, we train each model for 3 epochs with a batch size of $16$. This gives us $86.6\%$ and $82.0\%$ dense accuracy on RACE (H) and RACE (M), matching the ones reported in Fairseq ($86.5\%$ and $81.3\%$). Models on WinoGrande are trained for $23,750$ steps with $2,735$ warmup steps and $32$ batch size, reaching a $76.3\%$ accuracy. 

\begin{figure}[t]
\vspace{-2mm}
\centering
    \resizebox{0.98\textwidth}{!}{\includegraphics[]{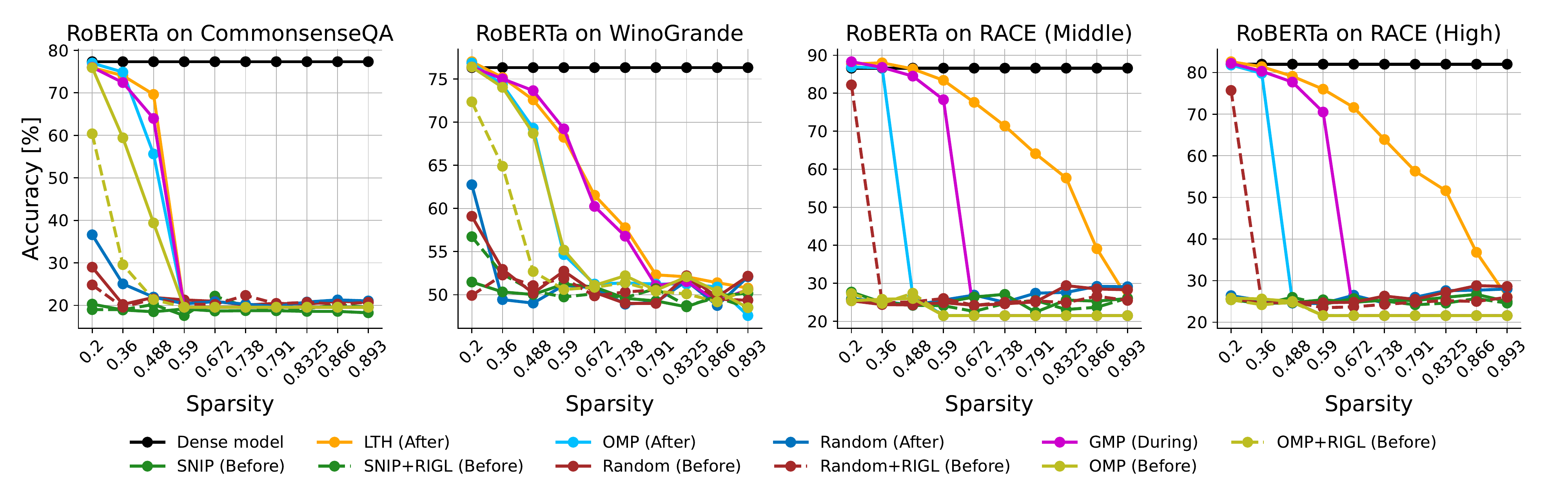}}
\vspace{-1.4em}
\caption{Commonsense reasoning performance of various sparse RoBERTa. }
\label{fig:common_sens}
\vspace{-1em}
\vspace{-2mm}
\end{figure}

\vspace{-2mm}
\paragraph{Results and Analyses.} 
The results of various sparse neural networks are demonstrated in Figure~\ref{fig:common_sens}. We summarize our main observations below:

\textit{\circled{1} All sparse algorithms seemingly fail to find matching SNNs, even at trivial sparsities such as 36\%.} While several methods maintain the dense performance at 20\% sparsity, their performance starts to drop significantly after that, and will undergo a catastrophic failure as the sparsity continues increasing. It is difficult even for the top-performing LTH to maintain the matching performance after the $2^{rd}$ IMP iteration. This is in stark contrast with the behavior of SNNs on the image classification task, where LTH can gracefully preserve the matching performance even at very extreme sparsities ($>$95\% on CIFAR-10/100~\citep{yin2022lottery} and $>$80\% on ImageNet~\citep{Renda2020Comparing}). 

\textit{\circled{2} The quality of SNNs on harder tasks suffers more from sparsity.} Models trained on the hardest task, CSQA, undergo a larger accuracy loss at the same sparsity than the other two datasets. For instance, all the SNNs on CSQA suffer from a catastrophic accuracy drop (up to  74\%) and become no better than the random prediction at 59\% sparsity. Meanwhile, when trained on WinoGrande and RACE at 59\% sparsity, two sparse algorithms (LTH and GMP) can maintain relatively good performance with a smaller performance loss (i.e., 3\% $\sim$ 10\%).

\looseness=-1 \textit{\circled{3} Post-training pruning consistently outperforms prior-training pruning.} LTH achieves the best performance across datasets, GMP performs well, and OMP (After) follows behind. However, prior-training pruning achieves worse performance. OMP (Before) performs closely behind OMP (After), whereas SNIP performs no better than the naive random pruning. After digging deeper into the case of SNIP, we find SNIP aggressively prunes the embedding layers to more than 99\% sparsity even with a mild overall sparsity of 20\%. Surprisingly, the leading dynamic sparsity approach RigL does not bring significant gains to these prior-training approaches, and sometimes even hurts the performance.

\vspace{-0.5em}
\subsection{Arithmetic Reasoning}
\vspace{-0.5em}
\begin{figure}[h]
\vspace{-1mm}
\centering
    \resizebox{0.9\textwidth}{!}{\includegraphics[]{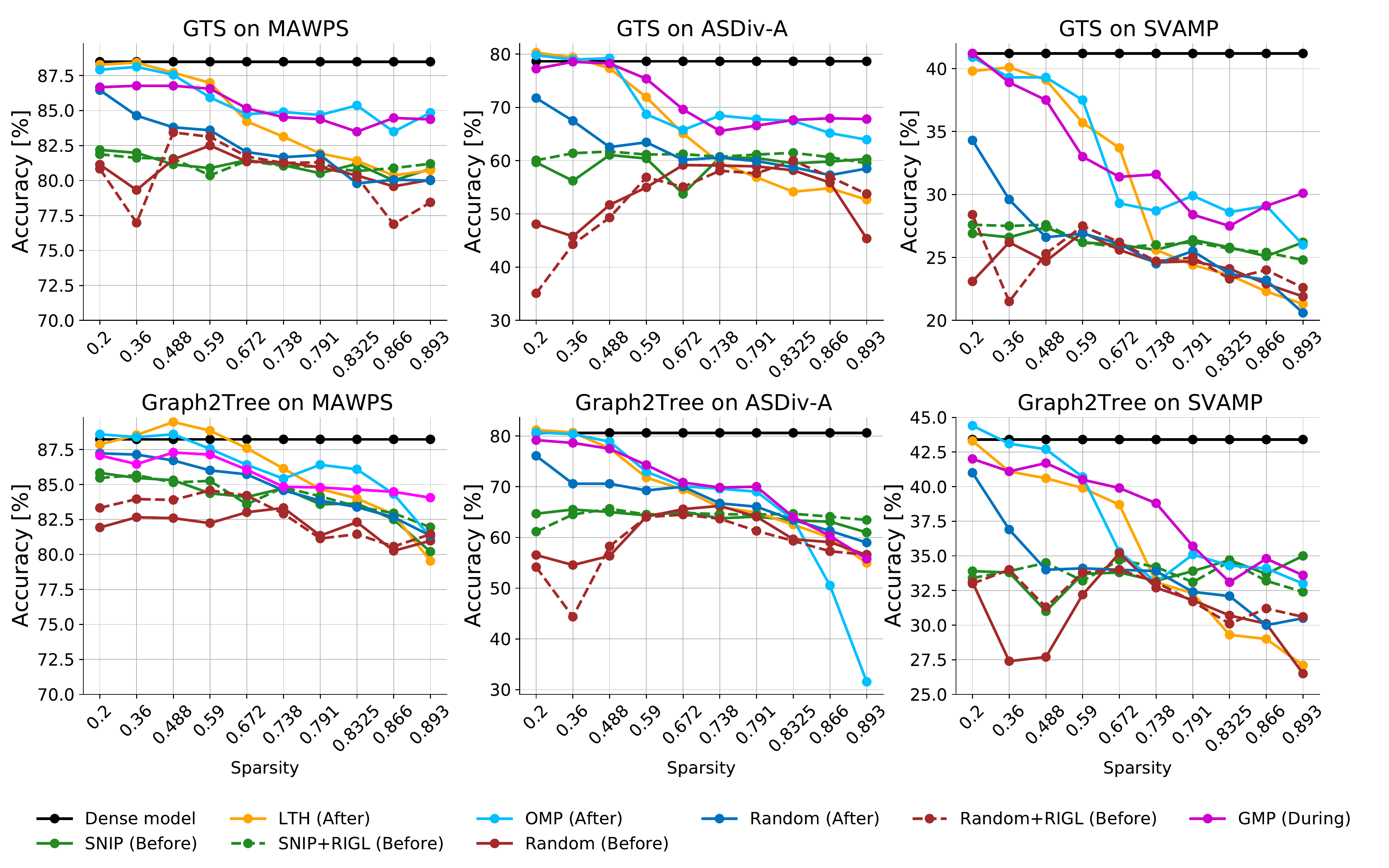}}
\vspace{-4mm}
\caption{Arithmetic reasoning performance of various sparse GTS and Graph2Tree.}
\label{fig:gts_math}
\vspace{-1em}
\vspace{-2mm}
\end{figure}

\looseness=-1 \paragraph{Implementation Details.} We follow SVAMP~\citep{patel2021nlp} and choose the two top-performing tree-based models for arithmetic reasoning: GTS~\citep{xie2019goal} and Graph2Tree~\citep{zhang2020graph}. Graph2Tree in general performs slightly better than GTS. GTS adopts LSTM to encode input sequences and a tree-based Decoder to generate questions. Graph2Tree uses a graph transformer to learn the latent quantity representations from data, and a tree structure decoder to generate a solution expression tree. We follow exactly the training settings of~\citet{patel2021nlp}. The embedding weights are inherited from THE pre-trained RoBERTa. All models are trained with Adam for 50 epochs. On MAWPS and ASDiv-A, models are trained with the training data and then evaluated on 5-fold cross-validation based on pre-assigned splits. For SVAMP, we train the models on a combination of MAWPS and ASDiv-A and test them on SVAMP, following~\citet{patel2021nlp}.  

\vspace{-2mm}
\paragraph{Results and Analyses.}
\looseness=-1 The performance on arithmetic reasoning is reported in Figure~\ref{fig:gts_math}. The overall performance trend is very similar to the commonsense reasoning: SNNs can only match the dense performance when the sparsity level is lower than 48.8\% with the exception of Graph2Tree on the relatively simple MAWPS dataset whose failing sparsity is $59\%$; SNNs are prone to sacrifice more performance on harder datasets (i.e., ASDiv-A and SVAMP) than the easier MAWPS dataset; prior-training  methods perform no differently than random pruning. Moreover, we want to highlight that LTH surprisingly reaches lower accuracy than OMP and GMP at high sparsity levels, indicating that iterative magnitude pruning may not necessarily generalize better than on more complex tasks. Moreover, Magnitude Before Training (OMG (Before)) consistently causes severe layer collapse in the non-embedding layers, leading to zero accuracies. Since including the results of OMG (Before) will significantly dilute the performance difference of different sparsification methods, we choose to report it in Appendix~\ref{app:mathall_w_omp}. 
\vspace{-0.5em}
\subsection{Protein Thermal Stability Prediction}
\vspace{-0.5em}

\subsubsection{Sequence-Based Models} 

\vspace{-1mm}
\paragraph{Implementation Details.} We examine two classic sequence-based approaches in protein property prediction, \textit{i.e.}, TAPE~\citep{rao2019evaluating} and ESM-1B~\citep{rives2021biological}. For TAPE, we fine-tune it from the official pre-training~\citep{rao2019evaluating} for $4$ epochs with an initial learning rate of $1\times10^{-4}$ and a linear decay scheduler together with $100$ warm-up steps. As for ESM-1B~\citep{rives2021biological}, starting from the official pre-trained checkpoint, we fine-tune the backbone with a learning rate of $1\times10^{-6}$ and the linear classification head on top of the backbone with a learning rate of $2\times10^{-2}$. The learning rate schedulers used for both backbone and linear head are OneCycle~\citep{smith2019super} decay schedulers. The training batch size is $2$ for Meltome Atlas and $3$ for HotProtein (HP-S). Classification accuracy on test sets is collected to measure the model performance.

\begin{figure}[t]
\centering
\vspace{-2mm}
\resizebox{0.98\textwidth}{!}{\includegraphics[]{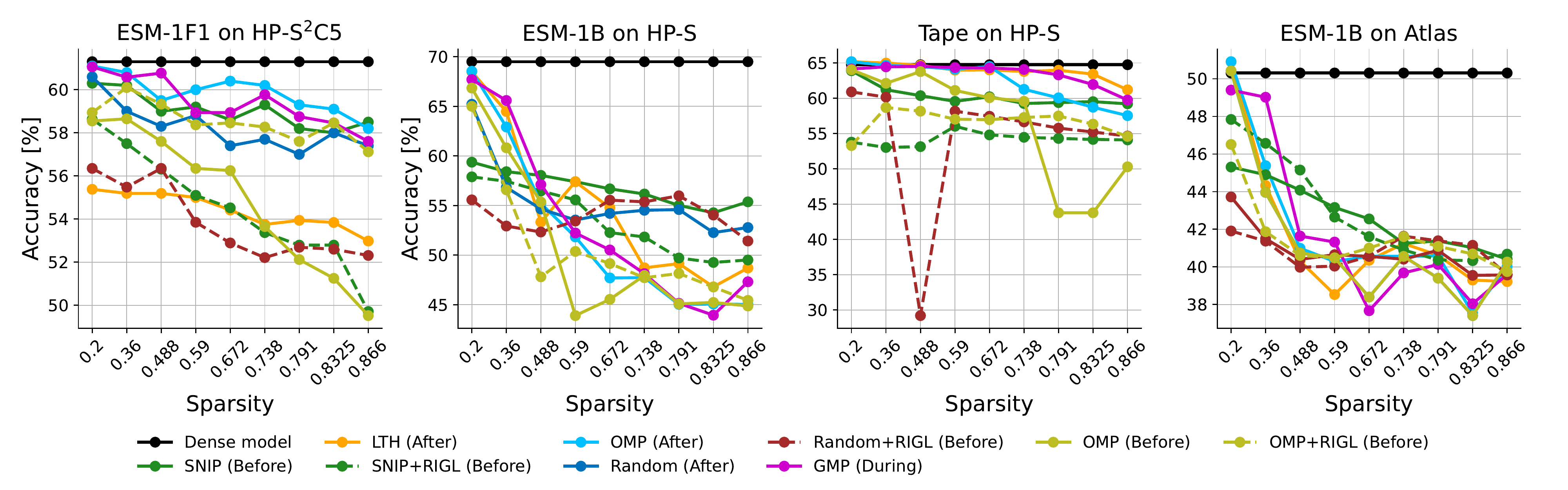}}
\vspace{-5mm}
\caption{Protein prediction performance of various sparse models.}
\label{fig:protein}
\vspace{-0.8em}
\vspace{-3mm}
\end{figure}

\vspace{-2mm}
\paragraph{Results and Analyses.} In this section, we examine diverse sparse neural networks of sequence-based models (\textit{i.e.}, transformers) on protein thermostability prediction benchmarks. ESM-1B~\citep{rives2021biological}, a SOTA approach in protein property modeling, is evaluated on both HotProtein (HP-S) and Meltome Atlas datasets. TAPE~\citep{rao2019evaluating} is a classic transformer-based model adopted on HotProtein (HP-S). Extensive results of both static and dynamic sparsifications are collected in Figure~\ref{fig:protein}. We observe that: \ding{182} For ESM-1B, all extracted sparse neural networks incur significant performance degradation whenever the sparsity level is larger than $20\%$. Note that here we only sparsify the fully connected layers in multi-head self-attentions \& feed-forward networks of each transformer layer, and leave all other modules unpruned. Even under this loose condition, ESM-1B still fails after pruning on both HP-S and Meltome Atlas, which indicates the low parameter redundancy in ESM-1B for modeling protein thermal stability. \ding{183} In general, static and dynamic pruning algorithms achieve similar performance with ESM-1B. SNIP (Before) and SNIP + RIGL (Before) deliver relatively better accuracies, especially for the high sparsity ($\ge 48.80\%$) subnetworks on HP-S. \ding{184} As for the worse backbone TAPE compared with ESM-1B, magnitude-based prunings like LTH (After), OMP (After), and GMP (During) show satisfactory results before $59\%$ sparsity.

\begin{wraptable}{r}{0.32\linewidth}
\centering
\vspace{-7mm}
\caption{\footnotesize OMP (after) pruning $5\%$ weights of ESM-1B on different modules with HP-S$^2$C2.}
\resizebox{0.7\linewidth}{!}{
\begin{tabular}{l|c}
\toprule
Pruned Modules & Accuracy ($\uparrow$)  \\
\midrule
None (Dense) & $94.68$\\ \midrule
Q, K, V, O, FFN & $92.19$ \\
Q, K, V & $92.55$ \\
Q, K & $93.62$ \\
Q, V & $93.62$ \\
K, V & $92.55$ \\
\bottomrule
\end{tabular}}
\vspace{-4mm}
\label{tab:protein_5}
\end{wraptable}
Furthermore, we conduct a more fine-grained pruning schedule to investigate the tolerance of ESM-1B against sparsification. In detail, we prune $5\%$ weights with OMP (after) on different modules in ESM-1B and present the obtained accuracies in Table~\ref{tab:protein_5}. Q, K, V, O, and FFN represent the fully connected layer in the query, key, value, \& output of the self-attention module and feed-forward networks, respectively. The results show that whatever modules we select, \textbf{$5\%$ sparsity} damages the ESM-1B performance of protein thermostability prediction on HP-S$^2$C2.

\vspace{-2mm}
\subsubsection{Structure-Based Models} 

\vspace{-1mm}
\paragraph{Implementation Details.} We further consider a representative structure-based algorithm for thermostability prediction, \textit{i.e.}, ESM-IF1~\citep{hsu2022learning}. Specifically, for ESM-IF1, we train the backbone and its linear head with learning rates of $1\times 10^{-4}$ and $2\times 10^{-2}$, respectively. A batch size of $4$ is used for both models on HotProtein (HP-S$^2$C$5$). Classification testing accuracy is reported to reflect the model performance. 

\vspace{-2mm}
\paragraph{Results and Analyses.} In this section, we study structure-based models and their sparse variants on HotProtein (HP-S$^2$C$5$). ESM-IF1~\citep{hsu2022learning}, a recent SOTA approach, is chosen for benchmarking. It takes the 3D structure of proteins as input and predicts its thermal stable temperature. As shown in Figure~\ref{fig:protein}, ESM-IF1 produces inferior sparse neural networks with all pruning mechanisms of both static and dynamic, where OMP (after) and GMP (During) present relatively better accuracies.

\vspace{-2mm}
\subsection{Multilingual Translation}
\vspace{-1mm}
\paragraph{Implementation Details.} We choose the official multilingual model mBART\footnote{\url{https://github.com/facebookresearch/fairseq}}~\citep{liu2020multilingual}, which was originally pre-trained on $25$ languages using masked language modeling (MLM), following the fine-tuning setting of~\citet{tang2020multilingual}. We first choose $10$ languages from the language pools used for MLM pre-training; create three sub-groups containing {$2$, $5$, $10$} languages; and fine-tune mBART on each sub-group, referring to $2$-to-$2$, $5$-to-$5$, and $10$-to-$10$ multilingual translation fine-tuning, respectively. During inference, we report the averaged BLEU~\citep{tang2020multilingual,liu2020multilingual} scores of bi-directional translation across 10 languages to measure the translation performance. Hence, the task difficulty monotonically decreases from $2$-to-$2$ to $5$-to-$5$, and to $10$-to-$10$ fine-tuning as more languages are involved during training.
The default fine-tuning configurations in~\citet{tang2020multilingual} are adopted for $40$K iterations with an Adam optimizer and a learning rate of $1\times 10^{-6}$. 

\vspace{-2mm}
\paragraph{Results and Analyses.} 

\begin{figure}[h]
\centering
\vspace{-0.5em}
    \resizebox{0.85\textwidth}{!}{\includegraphics[]{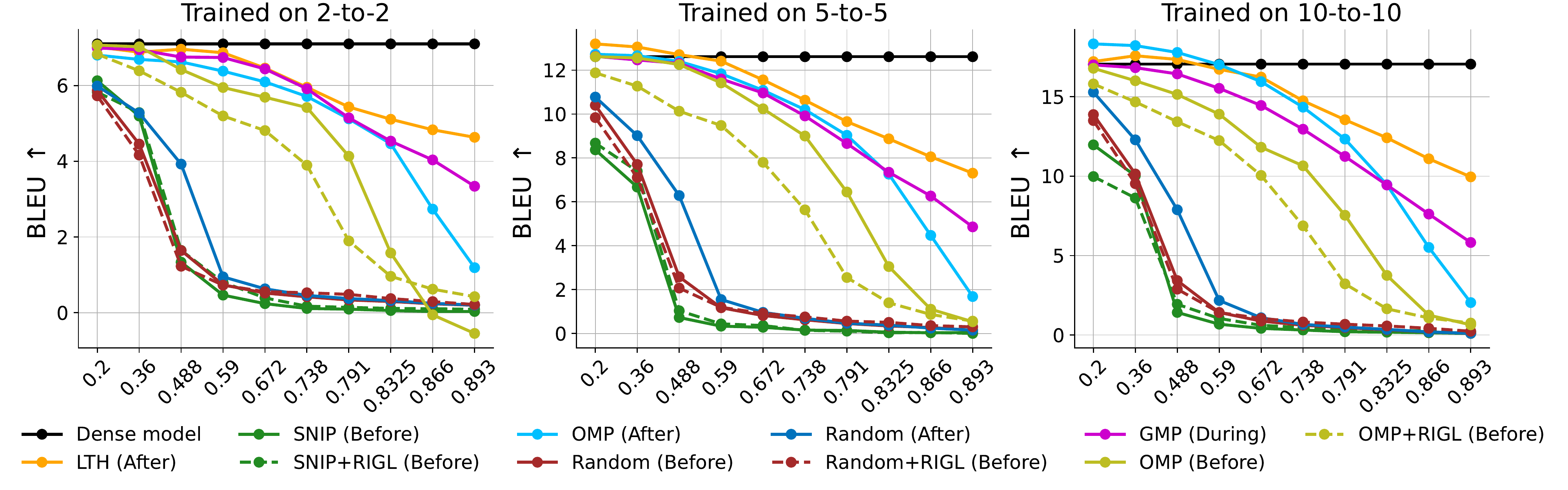}}
\vspace{-4mm}
\caption{Multilingual performance of various sparse mBART. All models are tested on 10-to-10 multilingual translation and the averaged BLEU are reported.}
\label{fig:multiling}
\vspace{-0.5em}
\vspace{-1mm}
\end{figure}

Intuitively, fewer languages involved during fine-tuning leads to a more difficult translation for all languages. As demonstrated in Figure~\ref{fig:multiling}, several consistent conclusions can be drawn: \ding{182} Besides OMP (After) and LTH (After), all other produced sparse subnetworks perform worse than the dense baseline when the sparsity is larger than $20\%$. The BLEU scores of OMP (After) and LTH (After) also decline and fail to match at $\ge 20\%$, $\ge 48.8\%$, $\ge 59\%$ sparsity levels for $2$-to-$2$, $5$-to-$5$, and $10$-to-$10$ fine-tuning, respectively. \ding{183} Magnitude-based sparsifications like OMP, LTH, and GMP are comparably robust across all three translation settings, while other pruning methods have negligible advantages compared to random pruning. \ding{184} While the overall tendency of SNNs is quite consistent across different tasks, the prunability of mBART increases as more languages are involved during fine-tuning. It seems that multilingual translation has already been a challenging task for pruning, and involving more languages in inference causes extra obstacles. This is why in the hardest scenario of fine-tuning on $2$-to-$2$ and evaluating with $10$ languages, all sparse subnetworks suffer from substantial performance degradation.

% \clearpage
\vspace{-0.6em}
\subsection{Why SNNs Fail on SMC-Bench}
\vspace{-0.6em}

We conduct a thorough investigation into the reasons why most SNNs struggle to perform on SMC-Bench. Our analysis identifies two possible causes for the failure: (1) the ``lazy regime'' in LLMs, and (2) the specific model components that are pruned. Based on these findings, we discover a set of stronger pruning recipes that alleviates (though still more or less suffering from) the failure on SMC-Bench, by breaking down the performance of the state-of-the-art BERT-pruning framework - oBERT~\citep{kurtic2022optimal} on SMC-Bench (note that most models evaluated in this paper are also BERT-based). 
Due to the limited space, we present our full investigation in Appendix~\ref{app:root_cause}, and briefly present our sanity check of layer collapse below.

% \textbf{Pruning embedding layers or not?} The pre-trained embeddings play a crucial role in the dense performance of SMC-Bench. For instance, the performance difference on ASDiv-A with and without RoBERTa's pre-trained embeddings can be up to 21.4\% as reported in~\citet{patel2021nlp}. To study whether the unsatisfactory performance of SNNs strongly correlates to the pruning of pre-trained embeddings, we conduct experiments where the embedding layers are kept dense. We can see from Figure~\ref{fig:noembed} while that pruning without embeddings unanimously brings large benefits to all approaches, it does not qualitatively change the fact that all these SOTA methods fail to match the dense performance at very naive sparsities. Moreover, since SNIP always aggressively prunes embedding layers, pruning without embeddings largely improves its performance (still behind 
% during- and post-training pruning). Therefore, we believe the sparsification of embedding layers is not the root cause for the failure of SNNs. \\
% \begin{wrapfigure}{r}{0.52\linewidth}
% \vspace{-4mm}
% \centering
% \includegraphics[width=1\linewidth]{images/CSQA_noembed.pdf}
% \vspace{-1.5em}
% \caption{Performance comparisons between sparsification with and without pruning embedding layers. ``w/o E'' refers to pruning without embedding layers.}
% \label{fig:noembed}
% \vspace{-1.2em}
% \end{wrapfigure}
\looseness=-1 \textbf{Does layer collapse occur unexpectedly on SMC-Bench?} Layer collapse is the most common cause that blocks the information flow (signal propagation) of sparse neural networks, leading to a catastrophic performance drop~\citep{tanaka2020pruning}. We plot the layerwise sparsity ratios of various sparse models in Appendix~\ref{app:layerwise_sr}. We do not observe severe layer collapse across methods except for SNIP which specifically removes nearly entire embedding layers. However, we do observe an unexpected phenomenon: \textit{layerwise sparsities of different magnitude-based pruning approaches (i.e., IMP, OMP, and GMP) are extremely similar,} all overlapped on one line, despite the significant performance gap among them (up to 42.3\%); small differences only start to appear until reaching very deep layers (e.g., classification heads) (see Appendix~\ref{app:layerwise_sr} for more details). This abnormal behavior is highly correlated with the ``lazy regime''~\citep{neyshabur2020being,malladi2022kernel} where the model stays in the same basin during fine-tuning with fairly small weight changes, and hence all magnitude-based pruning approaches, before, during, and after fine-tuning, tend to collapse to the same solution.

\vspace{-0.7em}
\section{Conclusion} 
\vspace{-0.7em}
Given the enormous breakthroughs and the fruitful results that sparse neural networks have achieved in numerous fields, it is necessary to rethink the sufficiency of current evaluation protocols and introduce more difficult and diverse benchmarks to explore the limitation of sparse neural networks. In this paper, we assemble a large-scale, challenging, and more diverse benchmark, SMC-Bench.
Through our thorough evaluation across various leading sparsifications, we confirm that SMC-Bench notably challenges the capability of most magnitude- or/and gradient-based sparse algorithms. We further dig deeper into the behavior of SNNs, and observe several surprising phenomena that are absent in the current evaluation. Our analysis points
out that such failure is intimately related to the ``lazy regime'', which leads us to a suite of strong pruning recipes that alleviates (yet still more or less suffering from) the failure on SMC-Bench. Our subsequent effort will focus on exploring stronger sparse training algorithms that can scale and generalize on SMC-Bench, and meanwhile will consider the training costs of different sparse algorithms for a more holistic picture of their efficiency benefits.

%In the hope to facilitate research endeavors to build next-generation sparse algorithms with the potential to generalize on complex and practical tasks, we open-source SMC-Bench at https://github.com/VITA-Group/SMC-Bench.   

\section*{Acknowledgement}

We thank Dan Alistarh and Eldar Kurtic for the extremely helpful discussions about the implementation details of oBERT as well as our benchmark's claims; and Zhangheng Li for helping run extra experiments with oBERT. S. Liu and Z. Wang are in part supported by the NSF AI Institute for Foundations of Machine Learning (IFML). Part of this work used the Dutch national e-infrastructure with the support of the SURF Cooperative using grant no. NWO2021.060, EINF-2694 and EINF-2943/L1.

\bibliography{iclr2023_conference}
\bibliographystyle{iclr2023_conference}

\clearpage
\appendix

\section{Summary of Tasks, Models, Datasets, and Training}
\label{app:summary}

We summarize the combinations of models and configurations that we used to evaluate SNNs on SMC-Bench.

\begin{table}[ht]
\centering

\caption{Summary of models and datasets that we used to evaluate on SMC-Bench.}
% \vspace{-0.2cm}
\label{tab:summary}
\resizebox{1.0\textwidth}{!}{
\begin{tabular}{l|ccc}
\toprule
\textbf{Task} & \textbf{Datasets} & \textbf{Models} & \textbf{Source} \\
\toprule

\multirow{3}{*}{\textbf{Commonsense Reasoning}} & CSQA & RoBERTa Large & Facebook AI Research Sequence-to-Sequence Toolkit~\citep{ott2019fairseq} \\
& WinoGrande & RoBERTa Large & Facebook AI Research Sequence-to-Sequence Toolkit~\citep{ott2019fairseq} \\
& RACE & RoBERTa Large & Facebook AI Research Sequence-to-Sequence Toolkit~\citep{ott2019fairseq} \\

\midrule
\multirow{6}{*}{\textbf{Arithmetic Reasoning}} & \multirow{2}{*}{MAVPS} & GTS  & GitHub Repository~\citep{patel2021nlp} \\

& & Graph2Tree  & GitHub Repository~\citep{patel2021nlp}\\
 \cmidrule(lr){2-4}
&\multirow{2}{*}{ASDiv-A} & GTS  & GitHub Repository~\citep{patel2021nlp} \\

& & Graph2Tree  & GitHub Repository~\citep{patel2021nlp}\\
 \cmidrule(lr){2-4}
 
&\multirow{2}{*}{SVAMP} & GTS  & GitHub Repository~\citep{patel2021nlp} \\

& & Graph2Tree  & GitHub Repository~\citep{patel2021nlp}\\
\midrule

\multirow{5}{*}{\textbf{Protein Thermostability Prediction}} & \multirow{2}{*}{HotProtein (HP-S)} & TAPE & GitHub Repository~\citep{rao2019evaluating}\\
& & ESM-1B & GitHub Repository~\citep{rives2021biological}\\ \cmidrule(lr){2-4}
& \multirow{1}{*}{HotProtein (HP-S$^2$C5)} & ESM-IF1 & GitHub Repository~\citep{rives2021biological}\\ \cmidrule(lr){2-4}
& \multirow{1}{*}{HotProtein (HP-S$^2$C2)} & ESM-1B & GitHub Repository~\citep{rives2021biological}\\ \cmidrule(lr){2-4}
& \multirow{1}{*}{Meltome Atlas} & ESM-1B & GitHub Repository~\citep{rives2021biological}\\
\midrule
\multirow{3}{*}{\textbf{Multilingual Translation}} & 2-to-2 & mBART & Facebook AI Research Sequence-to-Sequence Toolkit~\citep{ott2019fairseq} \\
& 5-to-5 & mBART & Facebook AI Research Sequence-to-Sequence Toolkit~\citep{ott2019fairseq} \\
& 10-to-10  & mBART & Facebook AI Research Sequence-to-Sequence Toolkit~\citep{ott2019fairseq} \\ 

\bottomrule
% \vspace{-1em}
\end{tabular}}
\end{table}

We strictly follow the training configurations reported in the original source to replicate the results of each task. The hyperparameters and configurations used for each model in this paper are shared below.
\subsection{Commonsense Reasoning}

\begin{table}[h]
	\centering
	\caption{\label{tab:hyperparams_cs}Hyperparameters and training configurations used for models on Commonsense Reasoning.}
	\resizebox{0.8\textwidth}{!}{
		\begin{tabular}{p{13em}P{8em}P{8em}P{8em}}
			\toprule
			\textbf{Models} &\textbf{RoBERTa}
			& \textbf{RoBERTa}  & \textbf{RoBERTa} \\ 
			\midrule
			Dataset &  CSQA & WinoGrande & RACE \\
			\midrule
			Pre-trained Models & RoBERTa & RoBERTa & RoBERTa  \\
			\midrule
			Hidden Size & [1024] & [1024]  & [1024] \\
			FFN Inner Hidden Size &  [4096] & [4096]  & [4096] \\
			Number of Layers & [24] & [24]  & [24]  \\
            
			Learning Rate & [1e-5] & [1e-5]  & [1e-5] \\
			
			Weight Decay & [0.01] & [0.01] & [0.01]  \\
			
			Batch Size & [16] & [32] & [16] \\
			
			Dropout & [0.1] & [0.1] & [0.1]\\
			
			Attention Dropout & [0.1] & [0.1] & [0.1] \\
			
			Clip Norm & [0.0] & [0.0] & [0.0] \\
			
			Adam $\epsilon$ & [1e-06] & [1e-06]  & [1e-06] \\
						
            Adam $\beta_1$ & [0.9] &  [0.9]  &  [0.9] \\
            
            Adam $\beta_1$ &  [0.98] & [0.98] & [0.98] \\
			\midrule
			
			\# Parameters & 355M & 355M &  355M \\
			
			Training Time & 3000 steps & 23750 steps &  3 epochs \\
			
			Wramup Time & 150 steps & 2375 steps & 500 steps\\
			\bottomrule
		\end{tabular} }
\end{table}

\clearpage
\vspace{-0.5em}
\subsection{Arithmetic Reasoning}
\vspace{-0.5em}

\begin{table}[h]
	\centering
	\caption{\label{tab:hyperparams_ar}Hyperparameters and training configurations used for models on Arithmetic Reasoning.}
	\resizebox{0.8\textwidth}{!}{
		\begin{tabular}{p{13em}P{12em}P{12em}}
			\toprule
			\textbf{Models} &\textbf{GTS}
			& \textbf{Graph2Tree} \\ 
			\midrule
			Dataset &  MAVPS, ASDiv-A, SVAMP &  MAVPS, ASDiv-A, SVAMP \\
			\midrule
			Pre-trained Embedding & RoBERTa & RoBERTa \\
			\midrule
			Embedding Size &  [768] &  [768] \\
			
			Hidden Size & [512] & [384] \\
			
			Number of Layers & [2] & [2]  \\
			
			Learning Rate & [1e-3] & [8e-4]  \\
			
			Weight Decay & [1e-5] & [1e-5] \\
			Embedding LR & [8e-6] & [1e-5] \\
			
			Batch Size & [4 (MAVPS, ASDiv-A), 8 (SVAMP)] & [4 (MAVPS, ASDiv-A), 8 (SVAMP)] \\
			
			Dropout & [0.5] & [0.5] \\
			
			Adam $\epsilon$ & [1e-08] & [1e-08]   \\
						
            Adam $\beta_1$ & [0.9] &  [0.9]  \\
            
            Adam $\beta_1$ &  [0.999] & [0.999]  \\
			\midrule
			
			\# Parameters & 140M & 143M \\
			
			Training Time & 50 epochs & 50 epochs \\
			
			\bottomrule
		\end{tabular}}

\end{table}

\vspace{-0.5em}
\subsection{Protein Thermostability Prediction}
\vspace{-0.5em}
\begin{table}[h]
	\centering
	\caption{\label{tab:hyperparams_pr}Hyperparameters and training configurations used for models on Protein Thermostability Prediction.}
	\resizebox{0.8\textwidth}{!}{
		\begin{tabular}{p{13em}P{8em}P{8em}P{8em}}
			\toprule
			\textbf{Models} & \textbf{TAPE}
			& \textbf{ESM-1B} & \textbf{ESM-IF1} \\ 
			\midrule
			Dataset & HP-S & HP-S$^2$C2, Meltome Atlas, HP-S &  HP-S$^2$C5 \\
			\midrule
		%	Pre-trained Embedding & &  \\
		%	Embedding Size &  [] &  [768] \\
			
			Hidden Size & [768] & [1280] & [512] \\
			
			Number of Layers & [12] & [33] & [20]  \\
			
			Learning Rate & [1e-4] & [2e-2 (head), 1e-6 (backbone)] & [2e-2 (head), 1e-4 (backbone)] \\
			
			Weight Decay & [1e-2] & [1e-2] & [5e-2] \\
			% Embedding LR & [8e-6] & [1e-5] \\
			
			Batch Size & [16] & [3,2,3] & [4]\\
			Attention Dropout & [0.1] & [0.0] & [0.1] \\
			Dropout & [0.1] & [0.0] & [0.1] \\
			
			Adam $\epsilon$ & [1e-08] & [1e-08] & [1e-08]  \\
						
            Adam $\beta_1$ & [0.9] & [0.9]  & [0.9]\\
            
            Adam $\beta_1$ &  [0.999] & [0.999] & [0.999] \\
			\midrule
			
			\# Parameters & 92M & 650M & 124M\\
			
			Training Time & 4 epochs & 4 epochs & 8 epochs \\
			
			\bottomrule
		\end{tabular}}

\end{table}

\vspace{-0.5em}
\subsection{Multilingual Translation}
\vspace{-0.5em}
\begin{table}[h]
	\centering
	\caption{\label{tab:hyperparams_mt}Hyperparameters and training configurations used for models on Multilingual Translation.}
	\resizebox{0.8\textwidth}{!}{
		\begin{tabular}{p{13em}P{8em}P{8em}P{8em}}
			\toprule
			\textbf{Models} &\textbf{mBART}
			& \textbf{mBART}  & \textbf{mBART} \\ 
			\midrule
			Dataset &  2-to-2 & 5-to-5 & 10-to-10 \\
			\midrule
			Pre-trained Models & mBART & mBART & mBART \\
			\midrule
			Hidden Size & [1024] & [1024]  & [1024] \\

			Number of Layers & [24] & [24]  & [24]  \\
            
			Learning Rate & [3e-5] & [3e-5]  & [3e-5] \\
			
			Weight Decay & [0.0] & [0.0] & [0.0]  \\
			
			Batch Size & [16] & [32] & [16] \\
			
			Dropout & [0.3] & [0.3] & [0.3]\\
			
			Attention Dropout & [0.1] & [0.1] & [0.1] \\
			
			Clip Norm & [0.0] & [0.0] & [0.0] \\
			
			Adam $\epsilon$ & [1e-06] & [1e-06]  & [1e-06] \\
			
            Adam $\beta_1$ & [0.9] &  [0.9]  &  [0.9] \\
            
            Adam $\beta_1$ &  [0.98] & [0.98] & [0.98] \\
			\midrule
			
			\# Parameters & 680M & 680M &  680M \\
			
			Training Time & 40,000 steps & 40,000 steps &  40,000 steps \\
			
			Wramup Time & 2,500 steps & 2,500 steps & 2,500 steps\\
			\bottomrule
		\end{tabular} }
\end{table}

\clearpage 

\section{Results of Arithmetic Reasoning with Magnitude before Training}
\label{app:mathall_w_omp}
In this appendix, we report the performance of SNNs on arithmetic reasoning including Magnitude before Training (OMP (Before)). We can clearly observe that the accuracy of magnitude pruning before training dramatically falls from 80\% to nearly 0\% when sparsity is larger than 36\%.
After checking the corresponding layerwise sparsity, we find that OMP (Before) completely removes all the weights from non-embedding and non-encoder layers, leading to severe layer collapse. 

\begin{figure}[h]
\centering
    \resizebox{1\textwidth}{!}{\includegraphics[]{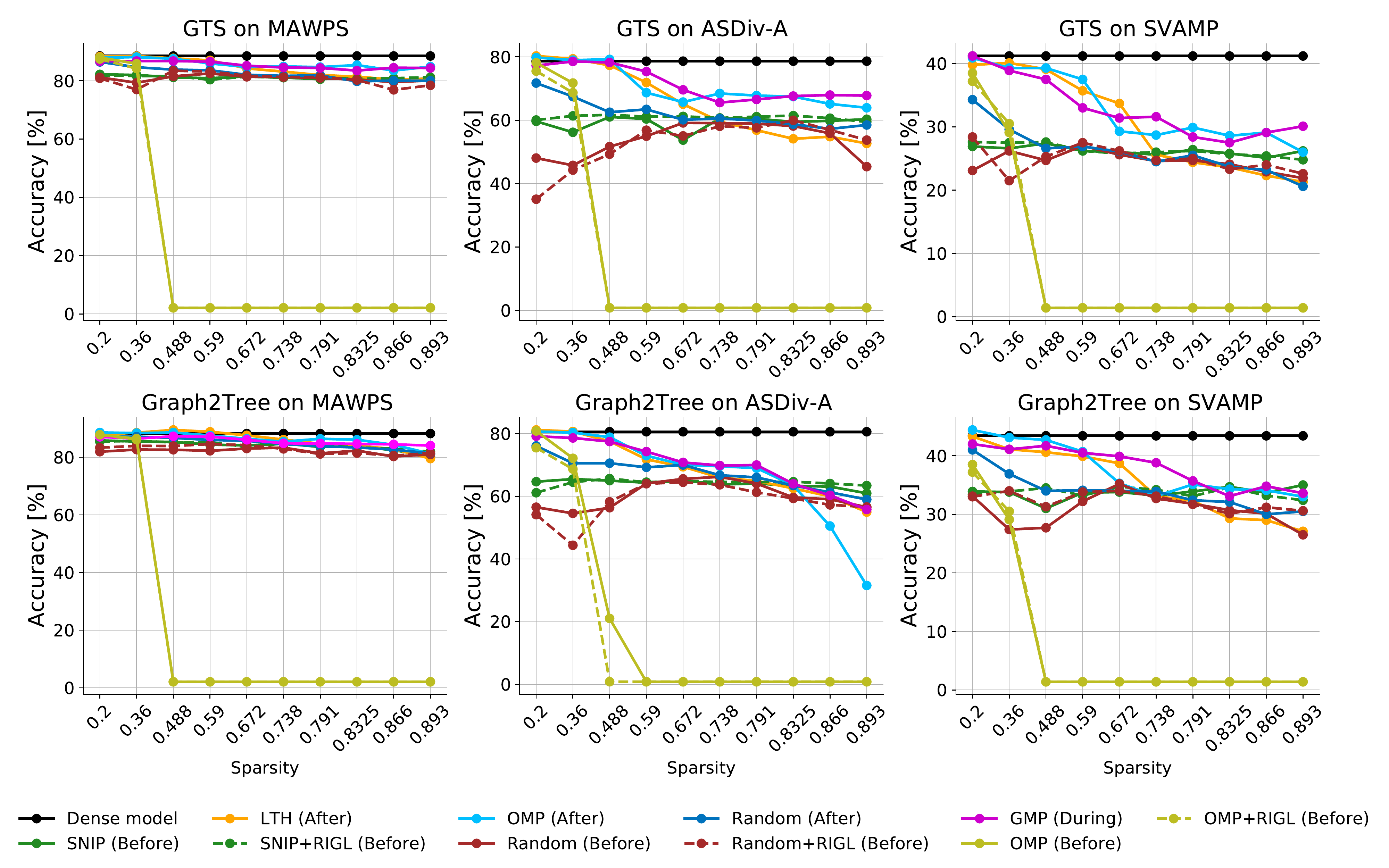}}

\caption{Arithmetic reasoning performance of various sparse GTS and Graph2Tree on MAWPS, ASDiv-A, and SVAMP.}
\label{fig:gts_math_w_omp}
\vspace{-1em}
\end{figure}

\clearpage
\section{An Investigation of Why SNNs Fail on SMC-Bench}
\label{app:root_cause}

In this section, we conduct a full investigation, attempting to open the box for the potential causes of SNN failures on SMC-Bench. Our analysis reveals two possible causes: (1) the ``lazy regime'' in fine-tuning LLMs, and (2) the model components to prune. Due to the ``lazy regime'' phenomenon~\citep{chizat2019lazy,malladi2022kernel}, commonly used pruning techniques that rely on magnitude and gradient can be very uninformative. Therefore, we turn to the latest strong second-order pruning framework - oBERT~\citep{kurtic2022optimal}, which utilizes inverse-Hessian approximations to guide pruning decisions. 
% In addition to the Hessian matrix, we also observe that oBERT  some other pruning techniques used in oBERT like iterative Learning Rate Rewinding~\citep{Renda2020Comparing}, Knowledge Distillation~\citep{hinton2015distilling}, tend to be beneficial in the ``lazy training'' regime and therefore matter for LLM pruning. 
We choose RoBERTa.large on CSQA to conduct this investigation. The roadmap of our full investigation is presented below. 
\begin{itemize}
    \item \textbf{Layerwise sparsity ratios} of various magnitude-based pruning methods are strikingly similar, suggesting that ``lazy regime'' may occur during fine-tuning. In this regime, the most common pruning criteria such as magnitude and gradient can be rather unreliable.
    \item \textbf{Second-order pruning} approaches like oBERT provide more faithful signals than magnitudes and gradients for LLM pruning, and hence achieve significantly higher accuracy at high sparsities. 
    % \item  \textbf{Which layer to prune matters}. We discover that excluding both the embeddings and classification heads from pruning helps prevent the accuracy collapse.
    
    % \item \textbf{Learning rate rewinding} (LRR) is crucial for LLM pruning, which significantly helps recover accuracy during fine-tuning, even through its final layer-wise sparsity ratios do not change compared to the one without LRR.
    % \item \textbf{Knowledge distillation} also brings benefits, but only marginal over LRR and the second-order criterion, in the context of LLMs pruning. It can be interpreted as externally encouraging the ``lazy'' model to be ``active'' by mimicking teachers' behavior  during fine-tuning, resulting in different layerwise sparsities as shown in Figure~\ref{fig:roadmap_sparsity}. 

\end{itemize}

\subsection{Layerwise Sparsity Ratios on SMC-Bench}
\label{app:layerwise_sr}

\begin{figure}[h]
\centering
    \resizebox{0.9\textwidth}{!}{\includegraphics[]{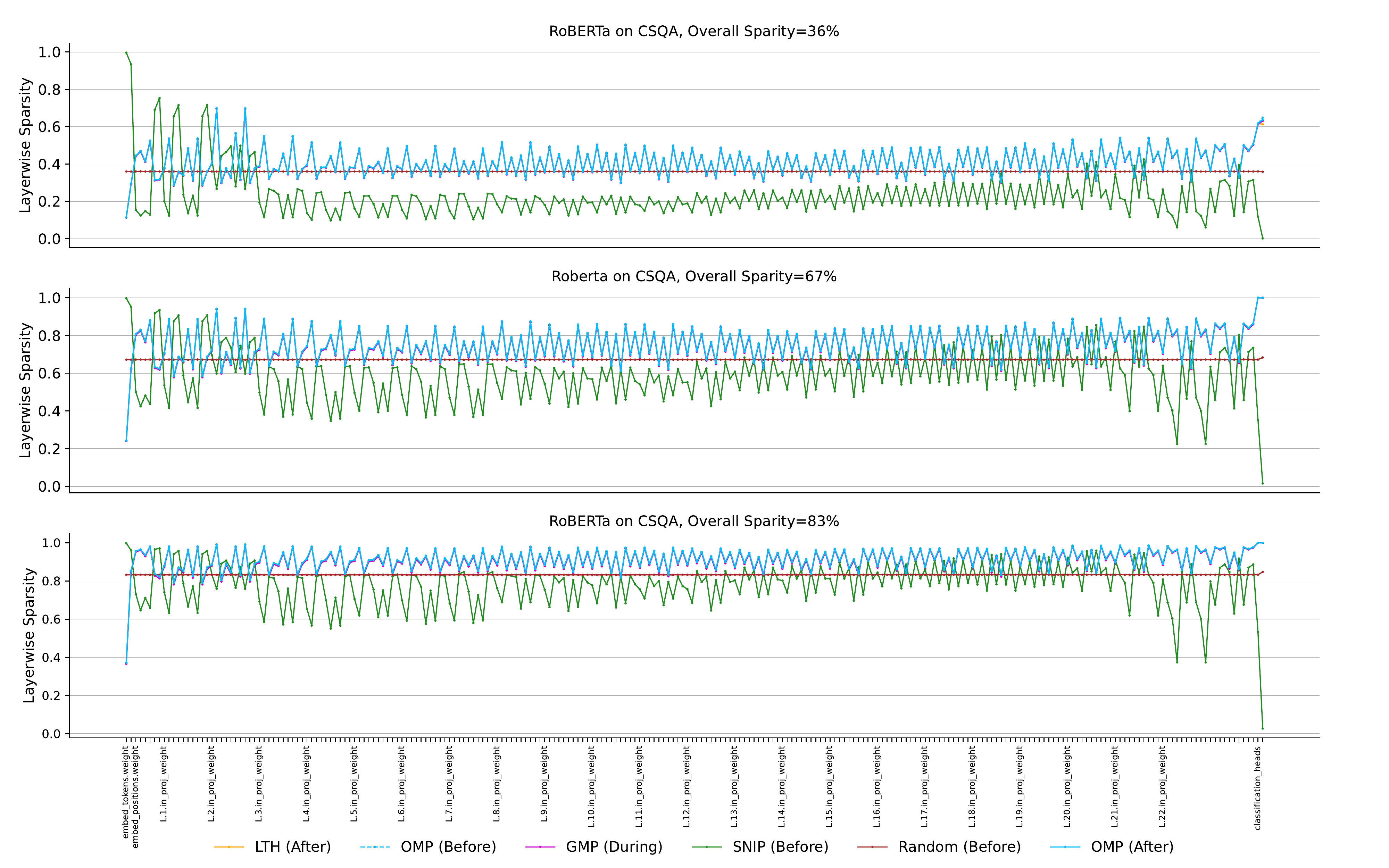}}

\caption{Layerwise sparsity of RoBERTa on CSQA at sparsity levels $ \in$ [36\%, 67\%, 83\%].}
\label{fig:layerwise_sparsity_csqa}
\vspace{-1em}
\end{figure}

\begin{figure}[h]
\centering
    \resizebox{0.9\textwidth}{!}{\includegraphics[]{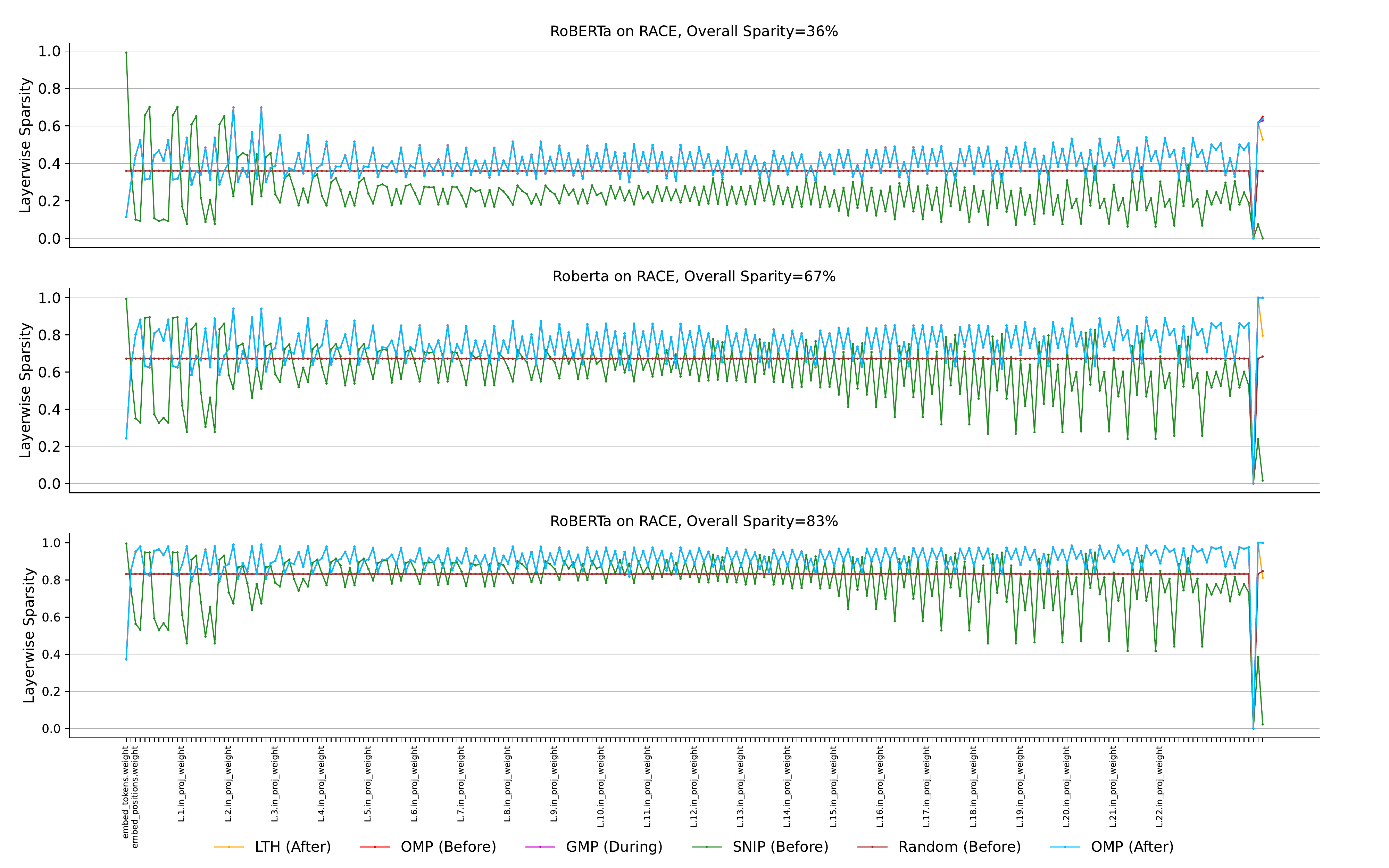}}

\caption{Layerwise sparsity of RoBERTa on RACE at sparsity levels $\in$ [36\%, 67\%, 83\%].}
\label{fig:layerwise_sparsity_race}
\vspace{-1em}
\end{figure}

\begin{figure}[h]
\centering
    \resizebox{0.9\textwidth}{!}{\includegraphics[]{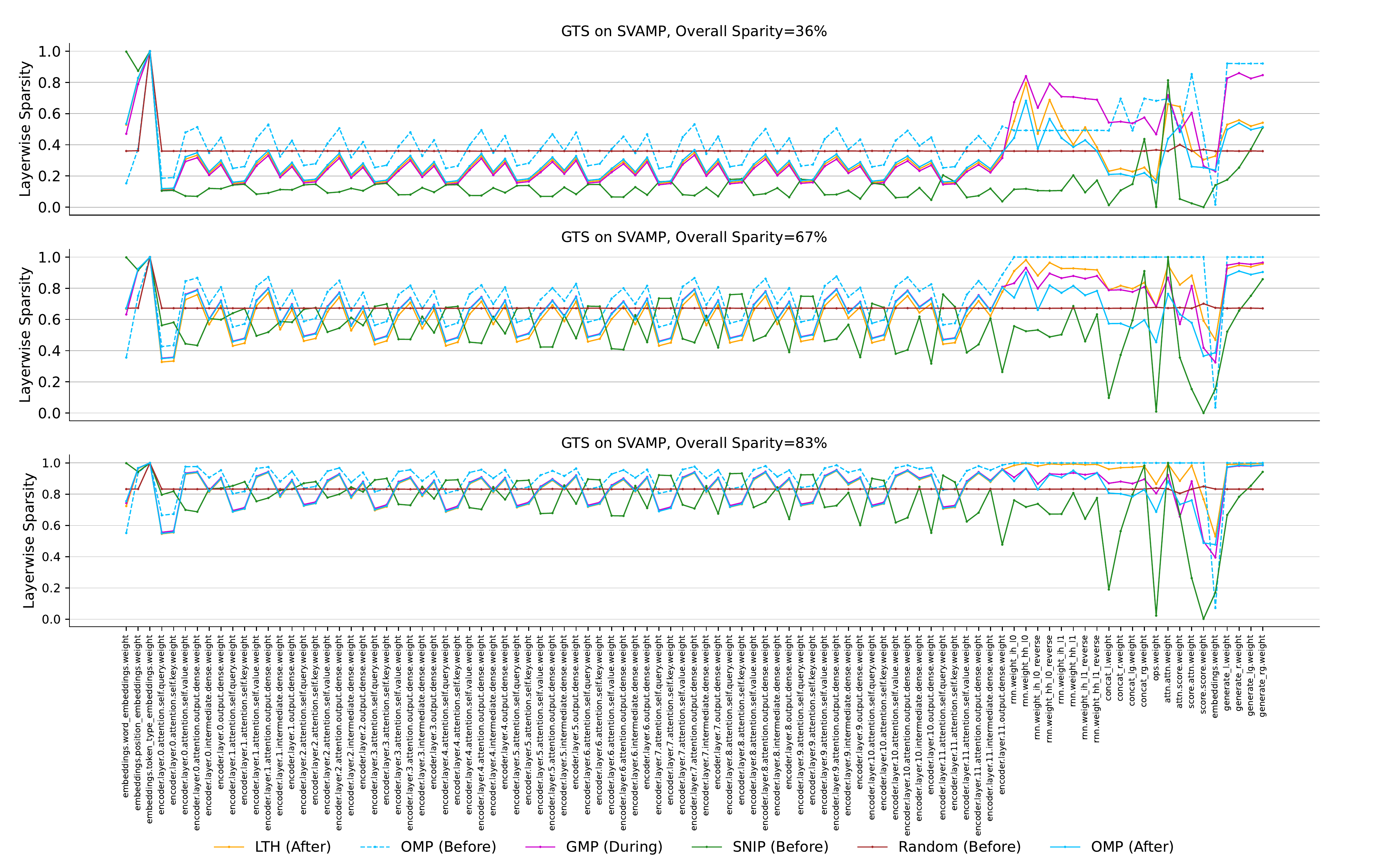}}

\caption{Layerwise sparsity of GTS on SVAMP at sparsity levels $\in$ [36\%, 67\%, 83\%].}
\label{fig:layerwise_sparsity_gts}
% \vspace{-1em}
\end{figure}

To check if severe layer collapse occurs on SMC-Bench, we plot the per-layer sparsity ratios discovered by various sparsification approaches at three sparsity levels: 36\%, 64\%, and 83\%. Layers are ordered from input to output on the X-axis. We respectively report the layerwise sparsity of commonsense reasoning with RoBERTa on CSQA and RACE in Figure~\ref{fig:layerwise_sparsity_csqa} and~\ref{fig:layerwise_sparsity_race}, and arithmetic reasoning with GTS on SVAMP in Figure~\ref{fig:layerwise_sparsity_gts}. We summarize our main findings here.

\ding{182} \textbf{Layerwise sparsities of magnitude-based pruning approaches are extremely similar.} IMP, OMP, and GMP that rely on weight magnitude for pruning share an extremely similar set of layerwise sparsities. Especially,  sparsity values of magnitude pruning on commonsense reasoning are completely identical, all overlapped on blue lines, except for the tiny difference in classification heads. This phenomenon indicates that weights of RoBERTa excluding classifiers remain rather stable during commonsense reasoning fine-tuning so that all the magnitude pruning variants (both before and after) discover the same sparsity pattern. The sparsity difference of arithmetic reasoning is more distinguishable than commonsense reasoning. Still, sparsities in the encoder (pre-trained RoBERTa) of IMP, GMP, and OMP (After) largely overlap. Until reaching the later layers, the sparsity ratios of different approaches start to be distinct. \ding{183} \textbf{SNIP tends to prune all the weights in embedding layers aggressively.} Even at the mild 36\% sparsity, SNIP prunes weights of embedding layers to 99.7\% sparsity, which may explain why SNIP struggles on SMC-Bench. \ding{184} \textbf{OMP (Before) suffers from layer collapse on arithmetic reasoning.} We empirically find that OMP (Before) leads to completely empty deep layers when sparsity is larger than 36\%, indicating the limitation of only considering pruning with the magnitude before fine-tuning or re-training.

The near-identical layerwise sparsity ratios across various magnitude-based  methods remind us of the “lazy training” regime~\citep{chizat2019lazy,malladi2022kernel} which was revealed to occur during the fine-tuning of LLMs. Under this regime,  weight changes during fine-tuning are negligible, hence non-informative and more ``noisy". Consequently, various magnitude-based pruning approaches, regardless of their timing, all tend to converge to the same solution.

% \begin{figure}[h]
% \centering
%     \resizebox{0.9\textwidth}{!}{\includegraphics[]{images/layerwise_sparsity_winogrande.pdf}}

% \caption{Layerwise sparsity of RoBERTa on WinoGrande at sparsity levels $ \in$ [36\%, 67\%, 83\%].}
% \label{fig:layerwise_sparsity_wino}
% \vspace{-1em}
% \end{figure}

\clearpage

\subsection{Evaluation of oBERT on SMC-Bench}
Based on our conjecture that magtitudes/gradients become unreliable for pruning LLMs, we hypothesize that the second-order pruning approaches with approximated Hessian matrix would be more accurate options. To verify our conjecture, we turn to the latest stronger second-order pruning framework, oBERT~\citep{kurtic2022optimal}. Specifically, we follow~\citet{kurtic2022optimal} and replace the magnitude pruning criterion of the LTH framework with the second-order oBERT criterion; adopt Learning Rate Rewinding (LRR)~\citep{Renda2020Comparing} and Knowledge Distillation (KD)~\citep{hinton2015distilling} during each pruning iteration; and keep the embeddings and classification heads dense. 

\begin{figure}[h]
\centering
    \resizebox{0.6\textwidth}{!}{\includegraphics[]{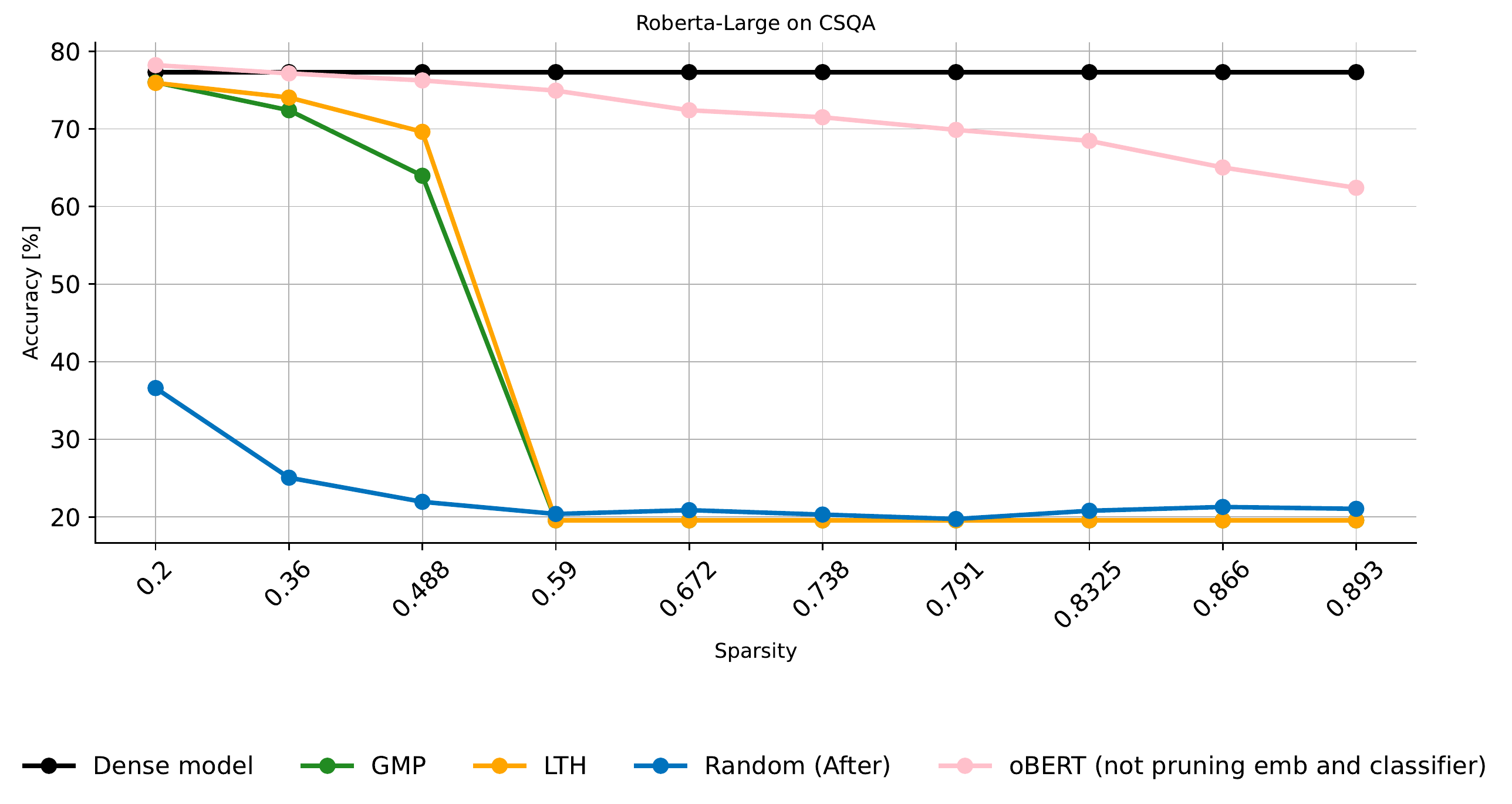}}
% \vspace{-1em}
\caption{A roadmap of accuracy recovering via a suite of stronger pruning recipes with RoBERTa-Large on CSQA.}
\label{fig:roadmap}
\vspace{-0.5em}
\end{figure}

Figure~\ref{fig:roadmap} support our hypothesis and demonstrates that oBERT notably outperforms the zero-order and first-order sparsification approaches, and substantially improves the accuracy to a competitive level. More importantly, oBERT produces a completely different layerwise sparsity pattern from magnitude-based pruning approaches, which is consistent with the patterns that are commonly observed in sparse computer vision models: deeper layers tend to have higher sparsities than lower layers~\citep{,evci2020rigging,kusupati2020soft,tanaka2020pruning,liu2021sparse}.

% We acknowledge that in addition to the second-order pruning matrix, the oBERT framework involves  three other pruning techniques that were not considered in the SMC-Bench evaluation protocol: selecting specific model components to prune, Learning Rate Rewinding, and Knowledge Distillation. To fully understand their effects on SMC-Bench, we decouple them from the oBERT pruner and report the corresponding performance in Figure~\ref{fig:roadmap}.

% \textbf{\ding{182} Keeping the embeddings and classification heads dense is crucial for  preventing performance collapse on SMC-Bench.} To demonstrate so, we try keeping  these layers dense while adopting  the most performant approach we have observed on SMC-Bench (i.e., LTH) on the remaining layers. Figure~\ref{fig:roadmap} illustrates that solely keeping embeddings dense  prevents LTH from collapsing until 79.1\% sparsity, and keeping both layers dense  manages to avoid random guessing at up to 89.3\% sparsity.

%Moreover, prior work~\citep{kurtic2022optimal} suggested that pruning these layers yields limited accelerations}

\begin{figure}[t]
\centering
    \resizebox{0.9\textwidth}{!}{\includegraphics[]{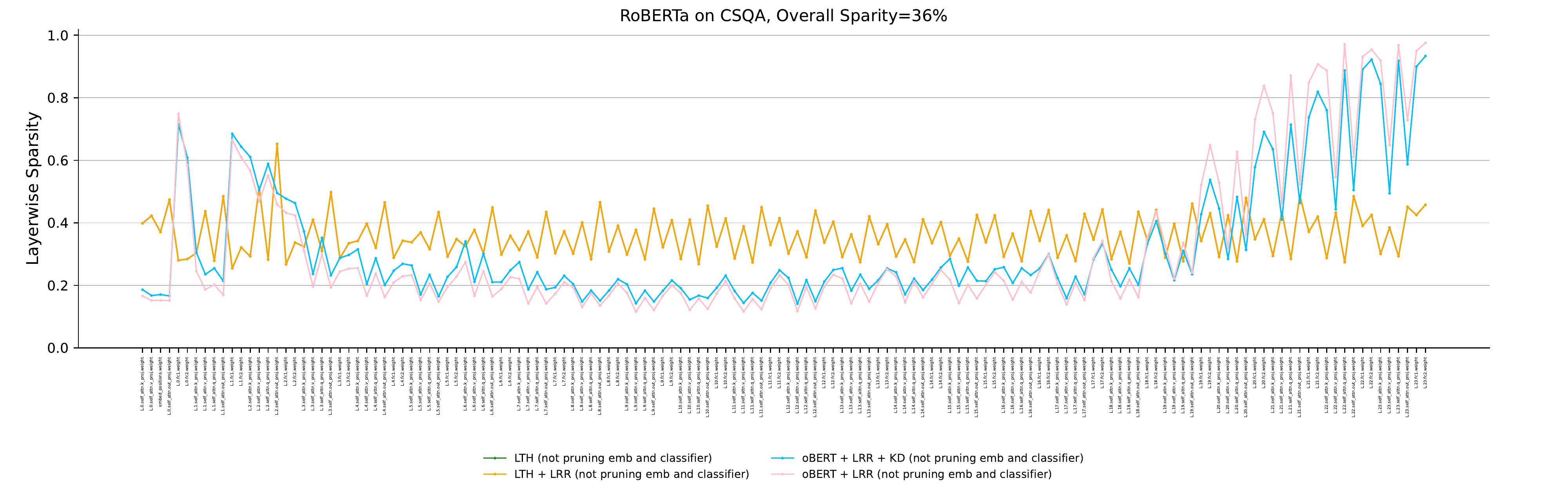}}
\vspace{-0.5em}
\caption{Layerwise sparsity comparison among LTH, LRR, oBERT, and KD with RoBERTa-Large on CSQA.}
\label{fig:roadmap_sparsity}
\vspace{-1em}
\end{figure}

So far, our investigation has discovered that the pruning recipe used in~\citep{kurtic2022optimal} can remarkably improve the LLM pruning performance on SMC-Bench. Nevertheless, such a well-tuned pruning recipe is both time- and resource-intensive (9$\times$ more fine-tuning time, besides Hessian matrix approximation); and even so, the strongest SNNs still fall short of their dense counterpart by around 10\% accuracy at sparsities between $60\% - 80\%$, in contrast to ``normal" SNNs that easily match their dense models on CIFAR, ImageNet, or GLUE. Therefore, the main claim of our paper still holds, that is, SMC-Bench indeed provides a new benchmark that is way more challenging for SOTA sparse algorithms, than existing testbeds.

% \vspace{-2em}
% \textcolor{blue}{\section{Limitation}
% \vspace{-0.5em}
% The main limitation of our paper is that we do not provide an apple-to-apple comparison among different sparse algorithms in terms of training/inference efficiency, but instead aim to highlight the limitations of the commonly-used evaluation protocols and assemble a more challenging one that the off-the-shelf SOTA sparsification algorithms fail to perform on. Hence, the training costs of different sparsification approaches evaluated in this paper vary a lot from prior-training to post-training. Practical designs would need to consider the total computational cost of training a sparse network but it is out of the scope of our current work.}

\clearpage
\section{Summary of Evaluation Tasks and Datasets in 100 Papers
\label{app:summary_100papers}}

\begin{table}[h]
\vspace{-1em}
\caption{Summary of Evaluation Tasks and Datasets Used in 100 Recent SNN Papers.}
\vskip -0.1 in
\begin{center}
\begin{tiny}
\begin{sc}
\resizebox{0.8\textwidth}{!}{
\begin{tabular}{c|c|cccc}
\toprule
\multirow{1}{*}{\textbf{Task}} & \multirow{1}{*}{\textbf{Total \#Paper}} & \textbf{Datasets} & \textbf{\#Paper} \\
\midrule
\multirow{9}{*}{Image Classification}&\multirow{9}{*}{82}
                                                          &ImageNet&62& \\
                                                        &&CIFAR-10&59&\\
                                                        &&CIFAR-100&37&\\
                                  &&MNIST& 26&\\ 
                                  &&Fashion MNIST&10&\\
                                  &&SVHN&4&\\
                                                        &&Birds-200&1&\\
                                                        &&Flowers-102&1&\\
                                  &&EMNIST &1&\\
\midrule
\multirow{9}{*}{NLP Task} &\multirow{9}{*}{16} &GLUE&9&\\
                                               & &SQUAD&4&\\
                                               & &WikiText-103&3&\\
                                               & &WMT&5&\\
                                                &&IMDB&1&\\
                                                &&AAN&1&\\
                                                &&LO&1&\\
                                                &&OpenWeb text&1&\\
                                                &&One Billion Word Benchmark&1&\\
                                                \midrule
\multirow{3}{*}{Face Recognition} &\multirow{3}{*}{3}&LFW&3&\\
                                                    &&Youtube Faces &2&\\
                                                    &&CASIA-WebFace&1&\\
                                                    \midrule
\multirow{2}{*}{Object Detection} &\multirow{2}{*}{3} & COCO dataset&2&\\
                                                    &&PASCAL-VOL-2007&1&\\
\midrule
\multirow{2}{*}{Speech Recognition}&\multirow{2}{*}{2}&Google-12&1&\\
                                                      &&TIMIT&1&\\
\midrule
\multirow{5}{*}{High-Resolution Reconstruction} &\multirow{5}{*}{2}&Set5&2&\\
                                                                    &&Set14&2&\\
                                                                    &&B100&2&\\
                                                                    &&Urban100&2&\\
                                                                    &&Manga109&2&\\
                                                                    \midrule
\multirow{3}{*}{Image Generation} &\multirow{3}{*}{2}&CIFAR-10&2&\\
                                                    &&ImageNet&1&\\
                                                    &&STL-10&1&\\
\midrule
\multirow{1}{*}{Human Activity Recognition} &\multirow{1}{*}{1}&HAR-2&1&\\
\midrule
\multirow{4}{*}{Microarray Classification} &\multirow{4}{*}{1}&Leukemia&1&\\
                                                                &&CLL-SUB-111&1&\\
                                                                &&SMK-CAN-18&1&\\
                                                                &&GLI-85&1&\\
\midrule
\multirow{1}{*}{Hand Gesture Reconstruction} &\multirow{1}{*}{1}& nvGesture &1&\\

\midrule
\multirow{1}{*}{Regression Task} &\multirow{1}{*}{1}&NYU Depth&1&\\
\midrule
\multirow{1}{*}{3D Object Part Segmentation} &\multirow{1}{*}{1}&ShapeNet&1&\\

\midrule
\multirow{4}{*}{RL task} &\multirow{4}{*}{1}&CartPole&1&\\
                                            &&Acrobot&1&\\
                                            &&MountainCar&1&\\
                                            &&Atari suite&1&\\
\midrule
\multirow{3}{*}{Vedio Deblurring} &\multirow{3}{*}{1}&DVD&1&\\
                                                    &&GOPRO&1&\\
                                                    &&Real Blurry Videos&1&\\

\midrule
\multirow{2}{*}{Vocabulary Speech Recognition} &\multirow{2}{*}{1}&VS&1&\\
                                                                    &&SWB&1&\\
% \midrule
% \multirow{1}{*}{Phonenme Recognition} &\multirow{1}{*}{1}&TIMIT&1&\\

\bottomrule
\end{tabular}}
\end{sc}
\end{tiny}
\end{center}
\label{tab:statistics}
\vskip -0.2in
\end{table}

\end{document}

%% file: math_commands.tex
\usepackage{amsmath,amsfonts,bm}

\def\eqref#1{equation~\ref{#1}}
\def\1{\bm{1}}

\DeclareMathAlphabet{\mathsfit}{\encodingdefault}{\sfdefault}{m}{sl}
\SetMathAlphabet{\mathsfit}{bold}{\encodingdefault}{\sfdefault}{bx}{n}